\documentclass[runningheads]{llncs}
\usepackage{graphicx}
\usepackage{makecell}
\usepackage[justification=centering]{caption}

\begin{document}
\title{Champion Team Paper: Dynamic Passing-Shooting Algorithm Based on CUDA of The RoboCup SSL 2019 Champion
}

\author{Zexi Chen$^1$, Haodong Zhang$^1$, Dashun Guo$^1$, Shenhan Jia$^1$, Xianze Fang$^1$,  Zheyuan Huang$^1$,  Yunkai Wang$^1$,  Peng Hu$^1$, Licheng Wen$^1$, Lingyun Chen$^1$, Zhengxi Li$^1$, and Rong Xiong$^1$}

\authorrunning{Z. Chen et al.}

\institute{Zhejiang University, Zheda Road No.38, Hangzhou, Zhejiang Province, P.R.China\\
\email{rxiong@zju.edu.cn}\\
\url{http://zjunlict.cn}
}
\maketitle
\begin{abstract}
ZJUNlict became the Small Size League Champion of RoboCup 2019 with $6$ victories and $1$ tie for their $7$ games. The overwhelming ability of ball-handling and passing allows ZJUNlict to greatly threaten its opponent and almost kept its goal clear without being threatened. This paper presents the core technology of its ball-handling and robot movement which consist of hardware optimization, dynamic passing and shooting strategy, and multi-agent cooperation and formation. We first describe the mechanical optimization on the placement of the capacitors, the redesign of the damping system of the dribbler and the electrical optimization on the replacement of the core chip. We then describe our passing point algorithm. The passing and shooting strategy can be separated into two different parts, where we search the passing point on \textit{SBIP-DPPS} and evaluate the point based on the ball model. The statements and the conclusion should be supported by the performances and log of games on Small Size League RoboCup 2019.

\keywords{Damping  \and STMicrocontrollers \and Off-the-ball Running \and Value-based Criteria \and CUDA Searching}
\end{abstract}

\section{Introduction}
ZJUNlict is a RoboCup Small Size League(SSL) team from Zhejiang University with a rich culture. We seek changes and upgrades every year from hardware to software, and try our best to fuse them together in order to form a better robot system. With a stable dribbler developed during 2017-2018, team ZJUNlict focused mostly on dynamic passing and shoot with the advantage of the stable dribbler in 2018-2019. In fact the algorithm helped us gain a ball possession rate of  $68.8\%$ during $7$ matches in RoboCup 2019.\\
To achieve the great possession rate, safe and accurate passing and shooting, our newly developed algorithm are developed into four parts:
\begin{enumerate}
	\item The passing point module calculates the feasibility of all passing points, filters out feasible points, and uses the evaluation function to find the best passing point.
	\item The running point module calculates the points where the offensive threat is high if our robots move there, to make our offense more aggressive. 
	\item The decision module decides when to pass and when to shoot based on the current situation to guarantee the success rate of the passing and shooting when the situation changes.
	\item The skill module helps our robots perform passing and shooting accurately.
\end{enumerate}

This paper focuses on how to achieve multi-robot cooperation. In Sects.2 and 3, we discuss our main optimization on hardware. In Sects.4 and 5, we discuss the passing strategy and the running point module respectively. In Sect.6, we analyze the performance of our algorithms at RoboCup 2019 with the log files recorded during the matches.

\section{Modification of Mechanical Structure of ZJUNlict}

\subsection{The position of two capacitors}
During a match of the Small Size League, robots could move as fast as $3.25 m/s$. In this case, the stability of the robot became very important, and this year, we focused on the center of the gravity with a goal of lower it. In fact, there are already many teams got there hands busy with lowering the center of the gravity, eg, team KIKS and team RoboDragon have their robot compacted to $135 mm$, and team TIGERs have their capacitor moved sideways instead of regularly laying upon the solenoid \cite{tiger}.

\begin{figure}
	\centering
	\includegraphics[width=0.5\textwidth]{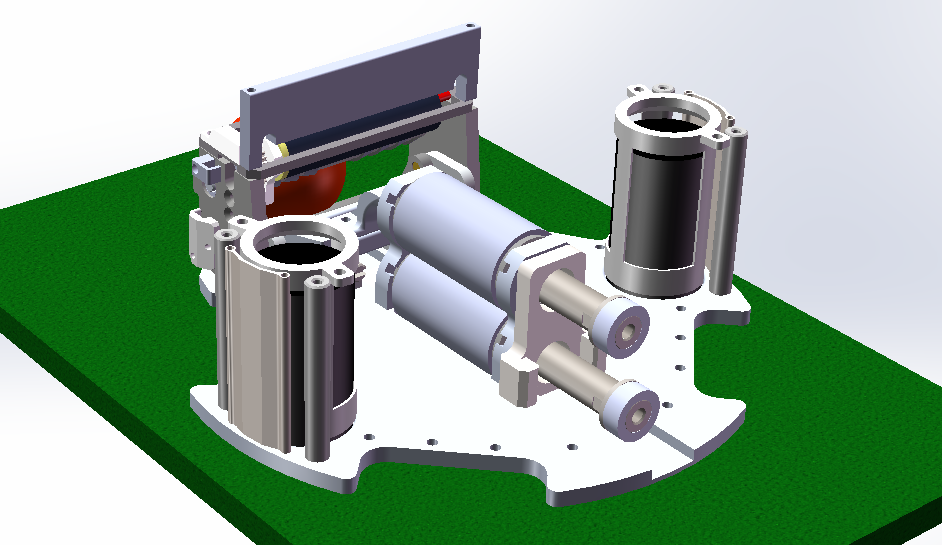}
	\caption{The new design of the capacitors}
	\label{FXZ-1}
\end{figure}

Thanks to the open source of team TIGERs \cite{tiger}, in this year's mechanical structure design, we moved the capacitor from the circuit board to the chassis. On the one hand, this lowers the center of gravity of the robot and makes the mechanical structure of the robot more compact, On the other hand, to give the upper board a larger space for future upgrades. The capacitor is fixed on the chassis via the 3D printed capacitor holder as shown in figure \ref{FXZ-1}, and in order to protect the capacitor from the impact that may be suffered on the field, we have added a metal protection board on the outside of the capacitor which made of 40Cr alloy steel with high strength.

\subsection{The structure of the dribbling system} 
The handling of the dribbling part has always been a part we are proud of, and it is also the key to our strong ball control ability. In last year's champion paper, we have completely described our design concept, that is, using a one-degree-of-freedom mouth structure, placing appropriate sponge pads on the rear and the lower part to form a nonlinear spring damping system. When a ball with certain speed hits the dribbler, the spring damping system can absorb the rebound force of the ball, and the dribbler uses a silica gel with a large friction force so that the ball can not be easily detached from the mouth.

\begin{figure}[htbp]
	\centering
	\begin{minipage}{6cm}
		\includegraphics[scale=0.22]{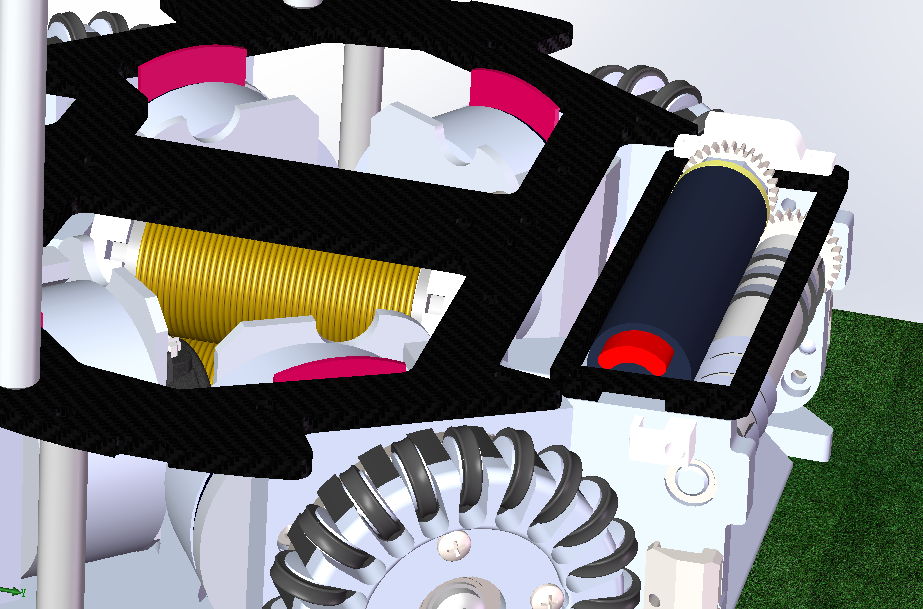}
		\caption{ZJUNlict 2018 mouth design}
		\label{FXZ-2}
	\end{minipage}%
	\begin{minipage}{6cm}
		\includegraphics[scale=0.22]{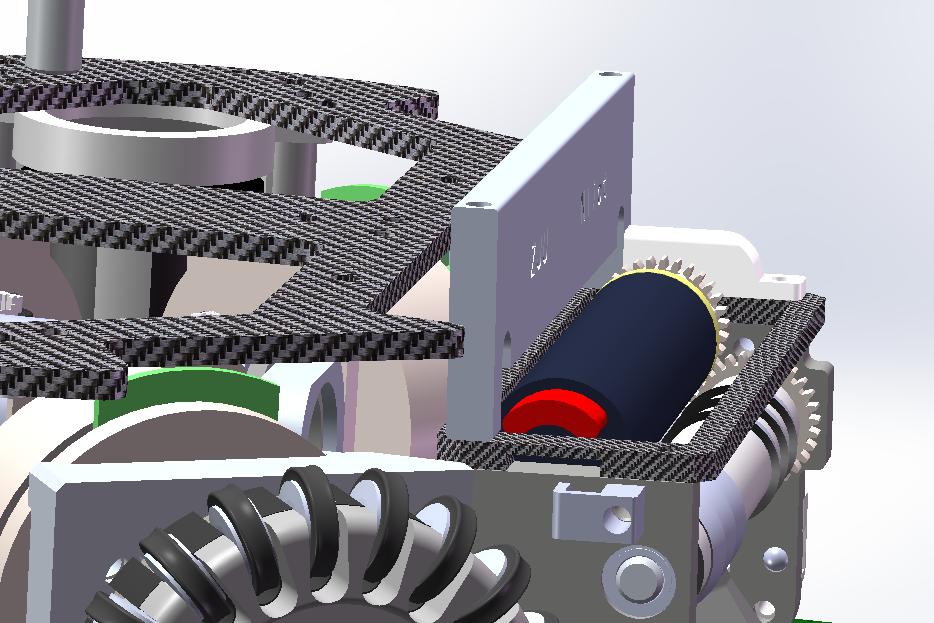}
		\centering\caption{ZJUNlict 2019 mouth design}
		\label{FXZ-3}
	\end{minipage}%
\end{figure}

The state of the sponge behind the mouth is critical to the performance of the dribbling system. In RoboCup 2018, there was a situation in which the sponge fell off, which had a great impact on the play of our game. In last year's design, as shown in figure \ref{FXZ-2}, we directly insert a sponge between the carbon plate at the mouth and the rear carbon plate. Under frequent and severe vibration, the sponge could easily to fall off\cite{champion2018}. In this case, we made some changes, a baffle is added between the dibbler and the rear carbon fiberboard, as shown in figure \ref{FXZ-3}, and the sponge is glued to the baffle plate, which made it hard for the sponge to fall off, therefore greatly reduce the vibration.

\section{Modification of Electronic Board}
In the past circuit design, we always thought that the board should be designed into multiple independent boards according to the function module so that if there is a problem, the whole board can be replaced. But then we gradually realized that instead of giving us convenience, it is unexpectedly complicated, on the one hand, we had to carry more spare boards, and on the other hand, it was not conducive to our maintenance.

For the new design, we only kept one motherboard and one booster board, which reduced the number of boards, making the circuit structure more compact and more convenient for maintenance. We also fully adopted ST's STM32H743ZI master chip, which has a clock speed of up to 480MHz and has a wealth of peripherals. The chip is responsible for signal processing, packet unpacking and packaging, and motor control.

Thanks to the open source of TIGERs again, we use Allergo's A3930 three-phase brushless motor control chip, simplifying the circuit design of the motor drive module on the motherboard. The biggest advancement in electronic this year was the completion of the stability test of the H743 version of the robot. In the case of all robots using the H743 chip, there was no robot failure caused by board damage during the game.

In addition, we replaced the motor encoder from the original 360 lines to the current 1000 lines. The reading mode has been changed from the original direct reading to the current differential mode reading.

\section{Passing and Shooting Strategy Based on Ball Model}
\subsection{Real-time Passing Power Calculation}
Passing power plays a key role in the passing process. For example, robot A wants to pass the ball to robot B. If the passing power is too small, the opponent will have plenty of time to intercept the ball. If the passing power is too large, robot B may fail to receive the ball in limited time. Therefore, it’s significant to calculate appropriate passing power.

\begin{figure}[h]
	\centering
	\includegraphics[width=0.5\textwidth]{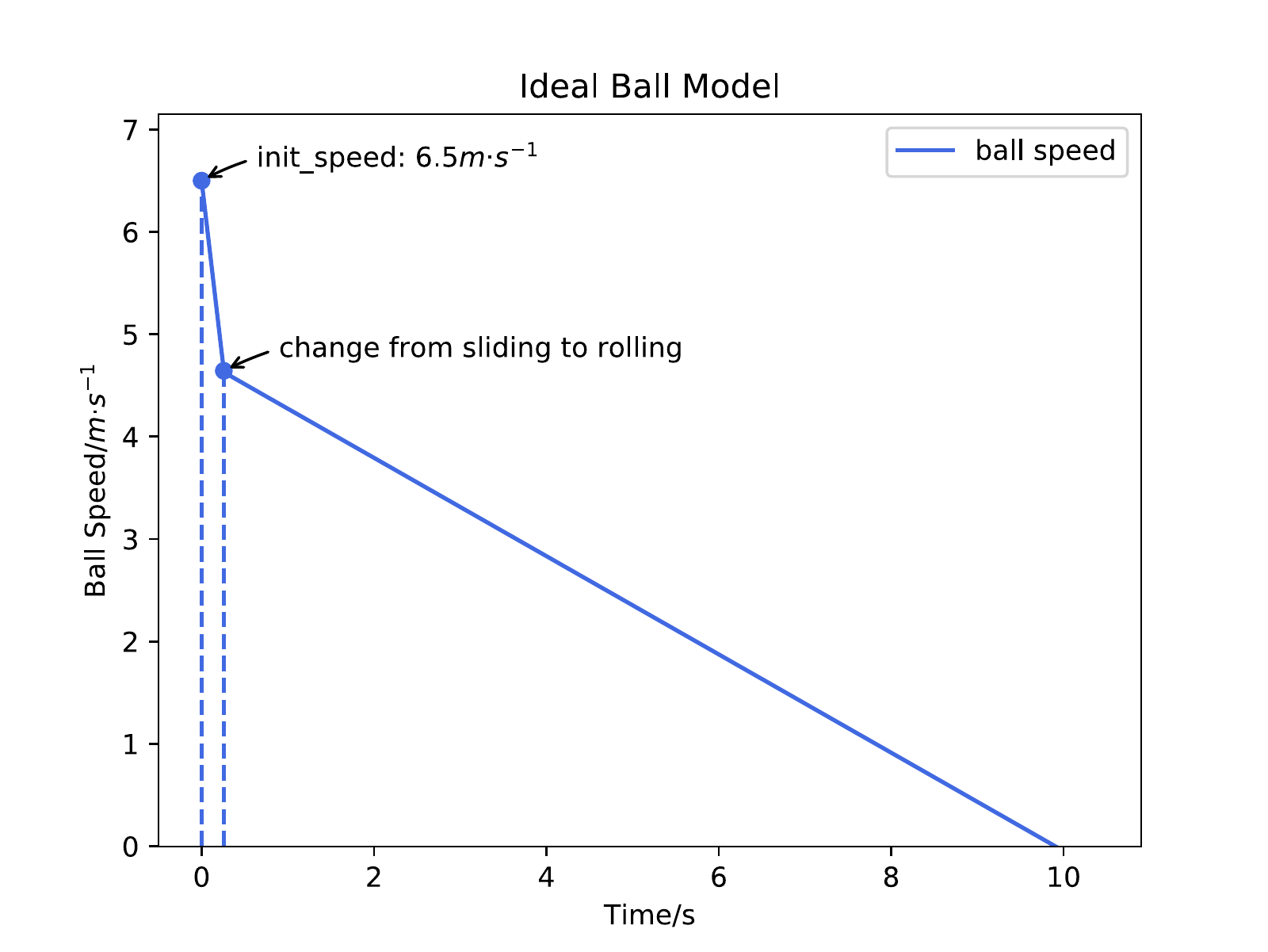}
	\caption{Ideal ball model}
	\label{DDQ}
\end{figure}

Suppose we know the location of robot A that holds the ball, its passing target point, and the position and speed information of robot B that is ready to receive the ball. We can accurately calculate the appropriate passing power based on the ball model shown in figure \ref{DDQ}. In the ideal ball model, after the ball is kicked out at a certain speed, the ball will first decelerate to $5/7$ of the initial speed with a large sliding acceleration, and then decelerate to $0$ with a small rolling acceleration. Based on this, we can use the passing time and the passing distance to calculate the passing power. Obviously, the passing distance is the distance between robot A and its passing target point. It’s very easy to calculate the Euclidean distance between these two points. Passing time consists of two parts: robot B’s arrival time and buffer time for adjustment after arrival. We calculate robot B’s arrival time using last year’s robot arrival time prediction algorithm. The buffer time is usually a constant (such as $0.3$ second). 

Since the acceleration in the first deceleration process is very large and the deceleration time is very short, we ignore the moving distance of the first deceleration process and simplify the calculation. Let d, t and a be the passing distance, time and rolling acceleration. Then, the velocity of the ball after the first deceleration and the passing power are given by the following:
\begin{equation}
v_1=((d+\frac{1}{2})at^2)/t  
\end{equation}

\begin{equation}
v_0=v_1/\frac{5}{7}
\end{equation}

According to the capabilities of the robots, we can limit the threshold of passing power and apply it to the calculated result.

\subsection{\textit{SBIP}-Based Dynamic Passing Points Searching (DPPS) Algorithm}
Passing is an important skill both offensively and defensively and the basic requirement for a successful passing process is that the ball can’t be intercepted by opponents. Theoretically, we can get all feasible passing points based on the \textit{SBIP(Search-Based Interception Prediction)} \cite{champion2018}\cite{etdp2019}. Assuming that one of our robots would pass the ball to another robot, it needs to ensure that the ball can’t be intercepted by opposite robots, so we need the SBIP algorithm to calculate interception time of all robots on the field and return only the feasible passing points. 

In order to improve the execution efficiency of the passing robot, we apply the searching process from the perspective of passing robot.

\begin{figure}[h]
	\centering
	\includegraphics[width=0.35\textwidth]{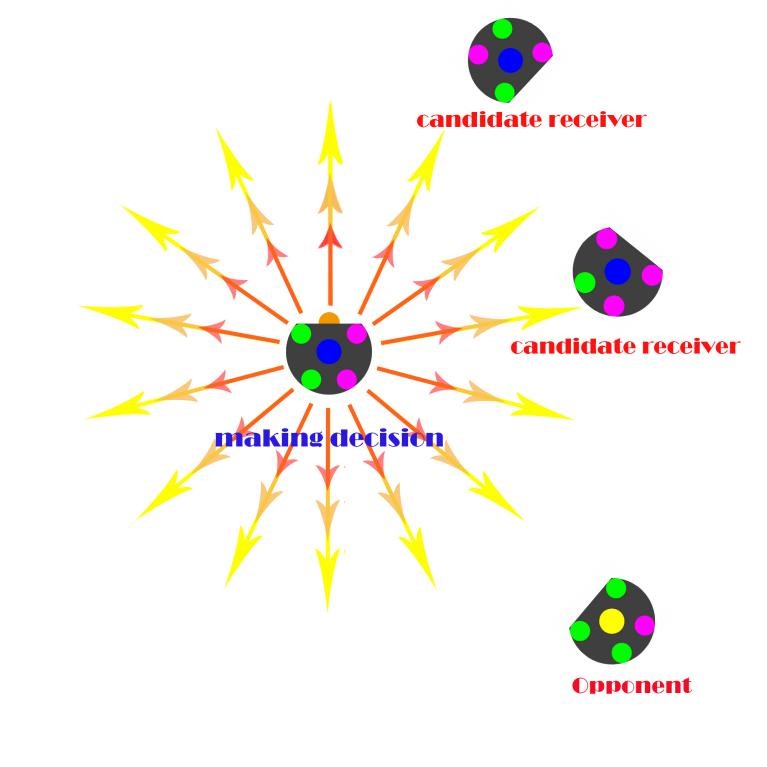}
	\caption{Dynamic passing points searching process}
	\label{JSH-1}
\end{figure}

As is shown in Figure \ref{JSH-1}, we traverse all the shooting power in all directions to apply the SBIP algorithm for all robots on the field. According to the interception time of both teammates and opponents under a specific passing power and direction, we can keep only the feasible passing directions and the corresponding passing power.

When considering that there is about $3$ degree’s error between the accurate orientation of the robot and the one obtained from the vision, we set the traversal interval of direction as $360/128$(about $2.8$) $degree$. And the shooting power, which can be considered as the speed of ball when the ball just kicked out, is divided equally into $64$ samples between $1 m/s$ and $6.5 m/s$, which means the shooting accuracy is about $0.34 m/s$. Because all combinations of passing directions and passing power should be considered, we need to apply SBIP algorithm for $262144$ times(we assume there are $16$ robots in each team, $32$ in the field), which is impossible to finish within about $13 ms$ by only serial computing. Fortunately, all of the $262144$ SBIPs are decoupled, so we can accelerate this process by GPU-based parallel computing technique\cite{CUDA-1}\cite{CUDA-2}\cite{CUDA-3}, and that’s why the numbers mentioned above are $128$, $64$ and $32$. 

\subsection{Value-based best pass strategy}\label{4.3}
After applying the DPPS algorithm, we can get all optional pass strategies. To evaluate them and choose the best pass strategy, we extract some important features $x_i (i = 1, 2, 3...n)$ and their weights $ (i = 1, 2, 3...n) $, at last, we get the scores of each pass strategy by calculating the weighted average of features selected equation \ref{sigema} \cite{soccer-1}\cite{soccer-2}
\begin{equation}
\label{sigema}
\sum_{i=1}^{n}\omega _i\cdot x_i
\end{equation}

\begin{figure}[htbp]
	\centering
	\begin{minipage}{6.5cm}
		\includegraphics[scale=0.3]{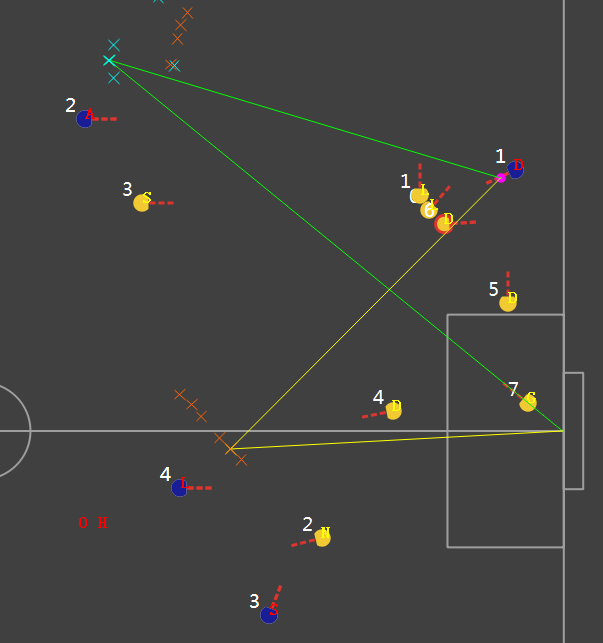}
		\label{JSH-2}\\
		\centering{a}
	\end{minipage}%
	\begin{minipage}{6.5cm}
		\includegraphics[scale=0.27]{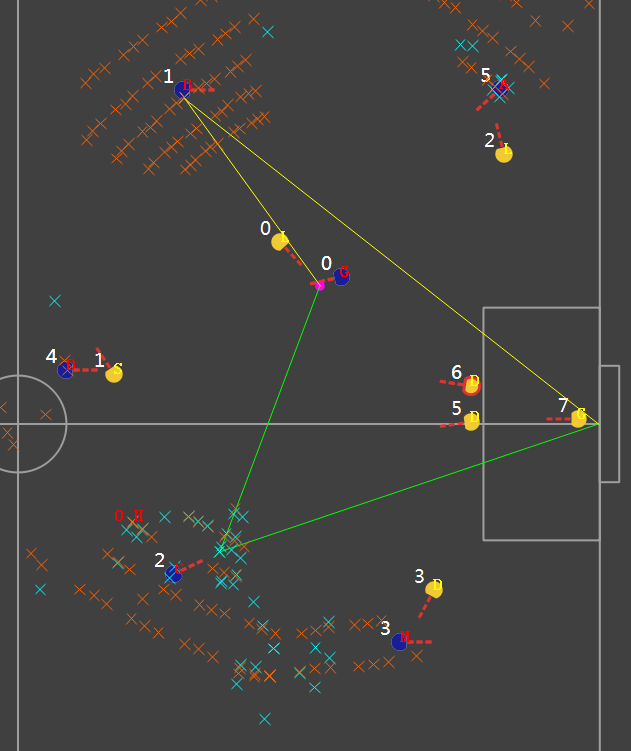}
		\label{JSH-3}\\
		\centering{b}
	\end{minipage}%
	\caption{Feasible pass points and best pass strategy}
	\label{JSH}
\end{figure}

For example, we chose the following features to evaluate pass strategies in RoboCup2019 Small Size League:

\begin{itemize}
	\item Interception time of teammates: close pass would reduce the risk of the ball being intercepted by opposite because of the ideal model.
	
	\item Shoot angle of the receiver’s position: this would make the teammate ready to receive the ball easier to shoot.
	
	\item Distance between passing point and the goal: if the receiver decides to shoot,
	short distance results in high speed when the ball is in the opponent's penalty
	area, which can improve the success rate of shooting.
	
	\item Refraction angle of shooting: the receiver can shoot as soon as it gets the ball if the refraction angle is small. The offensive tactics would be executed smoother when this feature is added.
	
	\item The time interval between the first teammate’s interception and the first opponent’s interception: if this number is very small, the passing strategy would be very likely to fail. So only when the delta-time is bigger than a threshold, the safety is guaranteed.
\end{itemize}
In order to facilitate adjustment of parameters, we normalize length and angle values by dividing their upper bound, while keeping the time values unchanged. 
After applying the DPPS algorithm, evaluating the passing points and choosing the best pass strategy, the results will be shown on the visualization software. In figure \ref{JSH}, the orange $\times$ is the feasible passing points by chipping and the cyan $\times$ is the feasible passing points by flat shot. The yellow line is the best chipping passing line, and the green line is the best flat shot passing line.

According to \textit{\textbf{a}} in figure \ref{JSH}, there are few feasible passing points when teammates are surrounded by opponents. And when the passing line is blocked by an opponent, there are only chipping passing points. According to \textbf{\textit{b}} in figure \ref{JSH}, the feasible passing points are intensive when there is no opponent marking any teammate.

\subsection{Shooting Decision Making}\label{shooting}
In the game of RoboCup SSL, deciding when to shoot is one of the most important decisions to make. Casual shots may lead to loss of possession, while too strict conditions will result in no shots and low offensive efficiency. Therefore, it is necessary to figure out the right way to decide when to shoot. We developed a fusion algorithm that combines the advantages of shot angle and interception prediction.

In order to ensure that there is enough space when shooting, we calculate the valid angle of the ball to the goal based on the position of the opponent’s robots. If the angle is too small, the ball is likely to be blocked by the opponent’s robots. So, we must ensure that the shot angle is greater than a certain threshold. However, there are certain shortcomings in the judgment based on the shot angle. For example, when our robot is far from the goal but the shot angle exceeds the threshold, our robot may decide to shoot. Because the distance from the goal is very far, the opponent’s robots will have enough time to intercept the ball. Such a shot is meaningless. In order to solve this problem, the shot decision combined with interception prediction is proposed. Similar to the evaluation when passing the ball, We calculate whether it will be intercepted during the process of shooting the ball to the goal. If it is not intercepted, it means that this shot is very likely to have a higher success rate. We use this fusion algorithm to avoid useless shots as much as possible and ensure that our shots have a higher success rate.

\subsection{Effective free kick strategy}
We generate an effective free kick strategy based on ball model catering to the new rules in 2019\cite{rules}. According to the new rules, the team awarded a free kick needs to place the ball and then starts the game in 5 seconds rather than 10 seconds before, which means we have less time to make decisions. This year we follow our one-step pass-and-shoot strategy, whereas we put the computation for best passing point into the process of ball placement. Based on the ball model and path planning, we can obtain the ball travel time $t_{p-ball}$ and the robot travel time $t_{p-robot}$ to reach the best passing point. Then we make a decision whether to make the robot reach the point or to kick the ball firstly so that the robot and the ball can reach the point simultaneously. 

Results in section \ref{result} show that this easy-executed strategy is the most effective strategy during the 2019 RoboCup Small Size League Competition.

\section{Off-the-ball Running}
\subsection{Formation}
\begin{figure}[htbp]
	\centering
	\begin{minipage}{5cm}
		\includegraphics[scale=0.22]{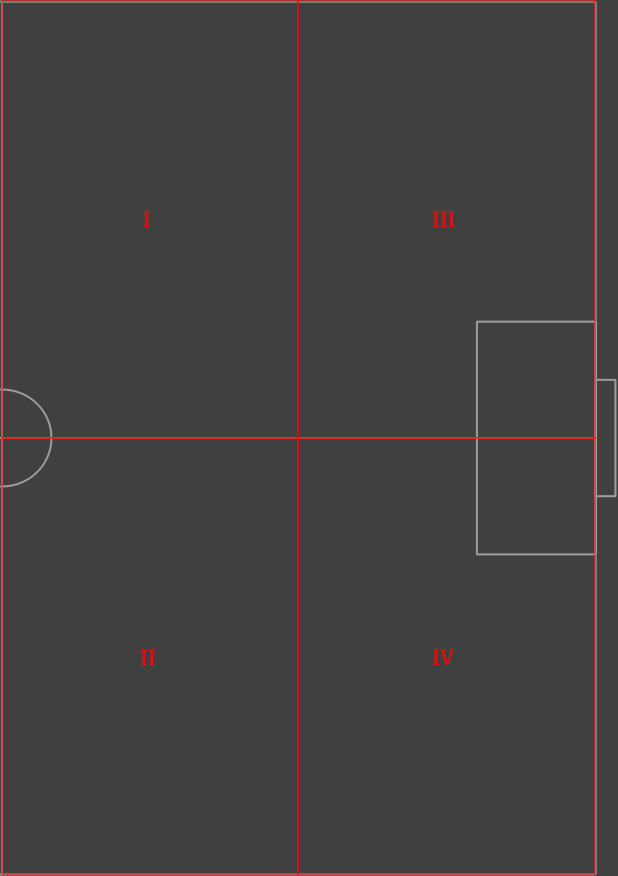}
		\caption{Four zones divided \protect\\by front field}
		\label{WZ-1}
	\end{minipage}%
	\begin{minipage}{5cm}
		\includegraphics[scale=0.65]{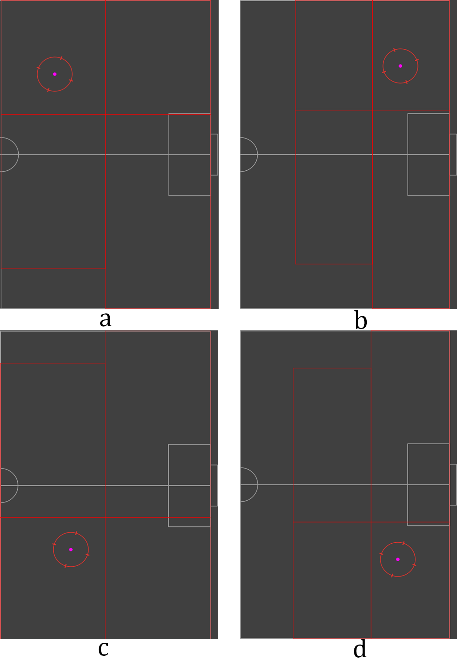}
		\caption{Dynamically changed zone\protect\\ according to the position of the ball}
		\label{WZ-2}
	\end{minipage}%
\end{figure}

As described in the past section, we can always get the best passing point in any situation, which means the more aggressiveness our robots show, the more aggressive the best passing point would be. There are two robots executing “pass-and-shot” task and the other robots supporting them\cite{robust}. We learned the strategy from the formation in traditional human soccer like “4-3-3 formation”and coordination via zones\cite{robot-soccer}. Since each team consists of at most $8$ robots in division A in 2019 season\cite{rules}, a similar way is dividing the front field into four zones and placing at most one robot in every part(figure \ref{WZ-1}). These zones will dynamically change according to the position of the ball(figure \ref{WZ-2}) to improve the rate of robot receiving the ball in it. Furthermore, we rasterize each zone with a fixed length (e.g. $0.1 m$) and evaluate each vertex of the small grids with our value-based criteria (to be described next). Then in each zone, we can obtain the best running point $x_R$ in a similar way described in section \ref{shooting}.

There are two special cases. First, we can’t guarantee that there are always $8$ robots for us on the field for yellow card and mechanical failure, which means at this time we can’t fill up each zone. Considering points in the zone III and IV have more aggressiveness than those in the zone I and II, at this time we prefer the best point in the zone III and IV. Secondly, the best passing point may be located in one of these zones. While trying to approach such a point, the robot may be possibly interrupted by the robot in this zone, so at this time, we will avoid choosing this zone.

\subsection{Value-based running point criteria}
\begin{figure}
	\centering
	\includegraphics[width=0.5\textwidth]{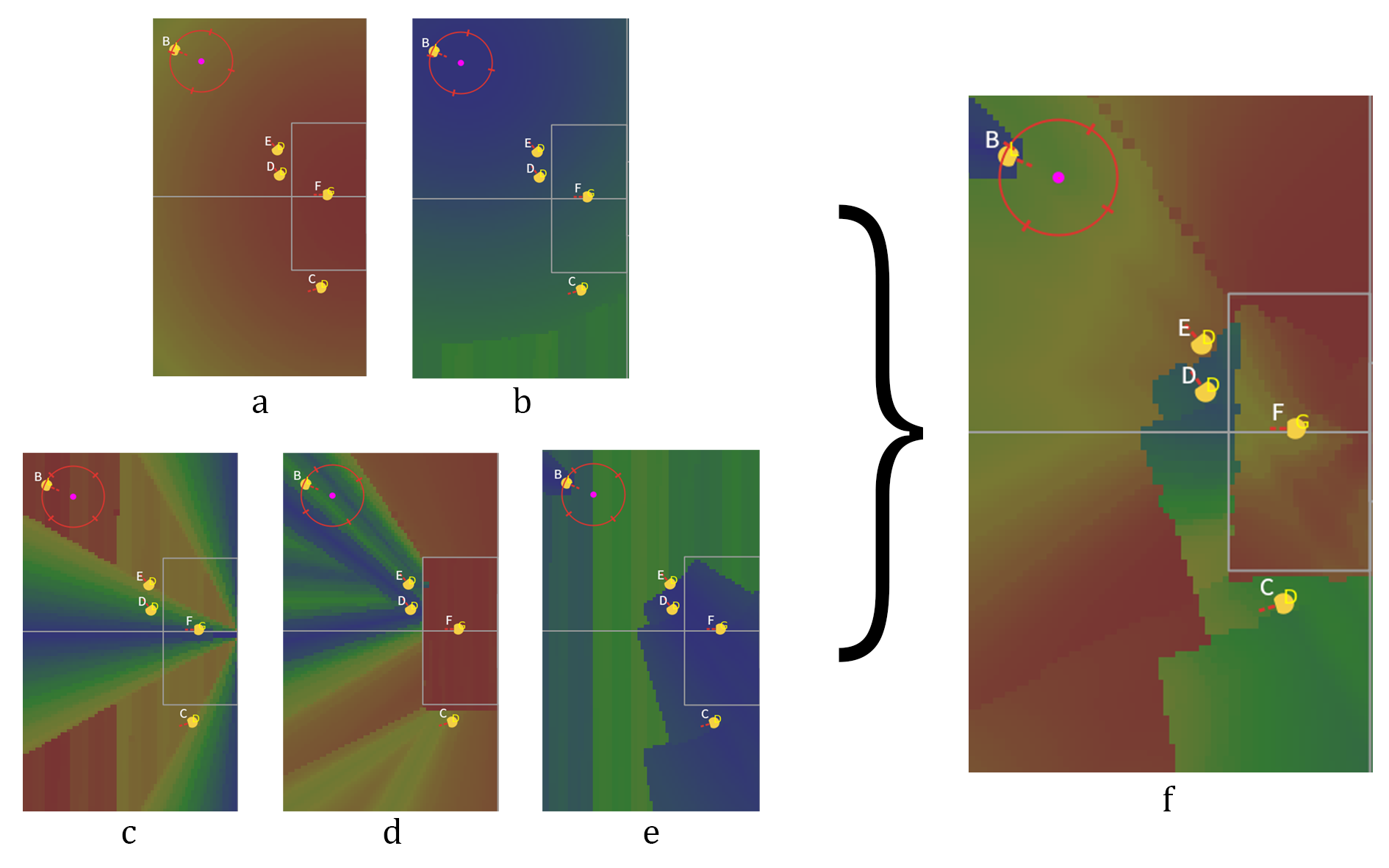}
	\caption{How Individual evaluation criterion affects the overall}
	\label{WZ-4}
\end{figure}

We adopt the similar approaches described in \ref{4.3} to evaluate and choose the best running point. There are five evaluation criteria $x_i (i=1,2,3...n)$ as follows. Figure \ref{WZ-4} shows how they work in common cases in order and with their weights $\omega_i (i=1,2,3...n)$ we can get the final result by equation \ref{equa} showed in f of figure  \ref{WZ-4} (red area means higher score while blue area means lower score).
\begin{equation}
\label{equa}
\sum_{i=1}^{n}\omega _i\cdot x_i
\end{equation}

\begin{figure}
	\centering
	\includegraphics[width=0.3\textwidth]{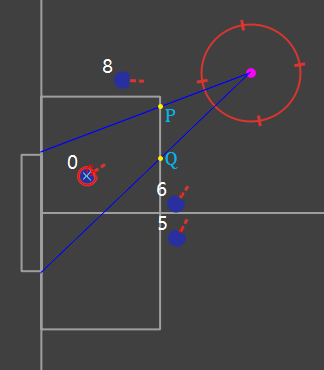}
	\caption{Method to get location P and Q}
	\label{WZ-3}
\end{figure}

\begin{itemize}
	\item \textbf{Distance to the opponent's goal.} It is obvious that the closer robots are to the opponent’s goal, the more likely robots are to score. 
	\item \textbf{Distance to the ball.} We find that when robots are too close to the ball, it is difficult to pass or break through opponent’s defense.
	\item\textbf{ Angle to the opponent's goal.} It doesn’t mean robot have the greater chance when facing the goal at 0 $degree$, instantly in some certain angle range.
	\item \textbf{Opponent’s guard time.} Guard plays an important role in the SSL game that preventing opponents from scoring around the penalty area, and each team have at least one guard on the field. Connect the point to be evaluated to the sides of the opponent’s goal, and hand defense area to $P$ and $Q$ (according to figure \ref{WZ-3}). Then we predict the total time opponent’s guard(s) spend arriving $P$ and $Q$. The point score is proportional to this time.
	\item \textbf{Avoid the opponent’s defense.} When our robot is further away from the ball than the opponent’s robot, we can conclude that the opponent’s robot will approach the ball before ours, and therefore we should prevent our robots being involved in this situation.
\end{itemize}

\subsection{Drag skill}
There is a common case that when our robot arrives at its destination and stops, it is easy to be marked by the opponent’s robot in the following time. We can call this opponent’s robot “defender”. To solve this problem, we developed a new “Drag” skill. First of all, the robot will judge if being marked, with the reversed strategy in \cite{etdp2019}. According to the coordinate information and equation(\ref{wz}) we can solve out the geometric relationship among our robot, defender and the ball, while they are clockwise with $Judge>0$ and counterclockwise with $Judge<0$. Then our robot will accelerate in the direction that is perpendicular to its connection to the ball. At this time, the defender will speed up together with our robot. Once the defender’s speed is greater than a certain value $v_{min}$, our robot will accelerate in the opposite direction. Thus there will be a huge speed difference between our robot and defender, which helps our robot distance defender and receive the ball safely. 

The application of this skill allows our robots to move off the opponent’s defense without losing its purpose, thus greatly improves our ball possession rate.
\begin{equation}
\label{wz} Judge=(x_{ball}-x_{me})(y_{opponent}-y_{me})-(x_{oppenent}-x_{me})(y_{ball}-y_{me})
\end{equation}

\section{Result}\label{result}
Our newly developed algorithms give us a huge advantage in the game. We won the championship with a record of six wins and one draw. Table 1 shows the offensive statistics during each game extracted from the official log.

The possession rate is calculated by comparing the interception time of both sides. If the interception time of one team is shorter, the ball is considered to be possessed by this team.

\begin{table}
	\caption{Statistics for each ZJUNlict game in RoboCup 2019}
	\begin{tabular}{|c|c|c|c|c|c|}
		\hline
		\textbf{Game} &  \textbf{\makecell[c]{Possession\\Rate(\%)}} & \textbf{\makecell[c]{Goals by\\Regular Gameplay}} & \textbf{\makecell[c]{Goals by\\Free Kick}} & \textbf{\makecell[c]{Goals by\\Penalty Kick}} & \textbf{Total Goals}\\
		\hline
		RR1 & 66.4 & 2 & 2 & 0 & 4\\
		\hline
		RR2 & 71.6 & 3 & 2 & 1 & 6\\
		\hline
		RR3 & 65.9 & 0 & 0 & 0 & 0\\
		\hline
		UR1 & -- & 2 & 1 & 1 & 4\\
		\hline
		UR2 & 68.2 & 1 & 0 & 1 & 2\\
		\hline
		UF & 69.2 & 1 & 1 & 0 & 2\\
		\hline
		GF & 71.4 & 1 & 0 & 0 & 1\\
		\hline
		Total & -- & 10 & 6 & 3 & 19\\
		\hline
		Average & 68.8 & 1.4 & 0.9 & 0.4 & 2.7\\
		\hline
	\end{tabular}

\end{table}

\subsection{Passing and Shooting Strategy Performance}
Our passing and shooting strategy has greatly improved our offensive efficiency resulting in 1.4 goals of regular gameplay per game. 52.6\% of the goals were scored from the regular gameplay. Furthermore, Our algorithms helped us achieve a 68.8\% possession rate per game.

\subsection{Free-kick Performance}
According to the game statistics, we scored an average of 0.9 goals of free-kick per game in seven games, while 0.4 goals for other teams in nineteen games. And goals we scored by free kick occupied 32\% of total goals (6 in 19), while 10\% for other teams (8 in 78). These statistics show we have the ability to adapt to new rules faster than other teams, and we have various approaches to score. 

\section{Conclusion}
In this paper, we have presented our main improvements on both hardware and software which played a key role in winning the championship. Our future work is to predict our opponent's actions on the field and adjust our strategy automatically. Improving our motion control to make our robots move faster, more stably and more accurately is also the main target next year.

\end{document}